\pdfoutput=1

\documentclass[11pt]{article}

\usepackage{emnlp2021}

\usepackage{times}
\usepackage{latexsym}

\usepackage[T1]{fontenc}

\usepackage[utf8]{inputenc}

\usepackage{microtype}

\usepackage{booktabs}
\usepackage{mathtools}
\usepackage{pifont}
\newcommand{\cmark}{\ding{51}}
\newcommand{\xmark}{\ding{55}}
\usepackage{amsmath}
\usepackage{amsfonts}

\usepackage[colorinlistoftodos,textsize=tiny]{todonotes}

\def\lowcomma{_{\textstyle,}}

\title{Expanding End-to-End Question Answering on Differentiable Knowledge Graphs with Intersection}

\author{Priyanka Sen \\
  Amazon Alexa AI \\
  Cambridge, UK \\
  \texttt{sepriyan@amazon.com} \\\And
  Amir Saffari \\
  Amazon Alexa AI \\
  Cambridge, UK \\
  \texttt{amsafari@amazon.com} \\\And
  Armin Oliya \\
  Amazon Alexa AI \\
  Cambridge, UK \\
  \texttt{aooliya@amazon.com} \\}

\begin{document}
\maketitle
\begin{abstract}
End-to-end question answering using a differentiable knowledge graph is a promising technique that requires only weak supervision, produces interpretable results, and is fully differentiable. Previous implementations of this technique \cite{cohen2020scalable} have focused on single-entity questions using a relation following operation. In this paper, we propose a model that explicitly handles multiple-entity questions by implementing a new \textit{intersection} operation, which identifies the shared elements between two sets of entities. We find that introducing intersection improves performance over a baseline model on two datasets, WebQuestionsSP (69.6\% to 73.3\% Hits@1) and ComplexWebQuestions (39.8\% to 48.7\% Hits@1), and in particular, improves performance on questions with multiple entities by over 14\% on WebQuestionsSP and by 19\% on ComplexWebQuestions.
\end{abstract}

\section{Introduction}

Knowledge graphs (KGs) are data structures that store facts in the form of relations between entities. Knowledge Graph-based Question Answering (KGQA) is the task of learning to answer questions by traversing facts in a knowledge graph. Traditional approaches to KGQA use semantic parsing to parse natural language to a logical query, such as SQL. Annotating these queries, however, can be expensive and require experts familiar with the query language and KG ontology. 

End-to-end question answering (E2EQA) models overcome this annotation bottleneck by requiring only weak supervision from question-answer pairs. These models learn to predict paths in a knowledge graph using only the answer as the training signal. In order to train an E2EQA model in a fully differentiable way, \newcite{cohen2020scalable} proposed differentiable knowledge graphs as a way to represent KGs as tensors and queries as differentiable mathematical operations.

\begin{figure}[ht]
\includegraphics[width=0.5\textwidth]{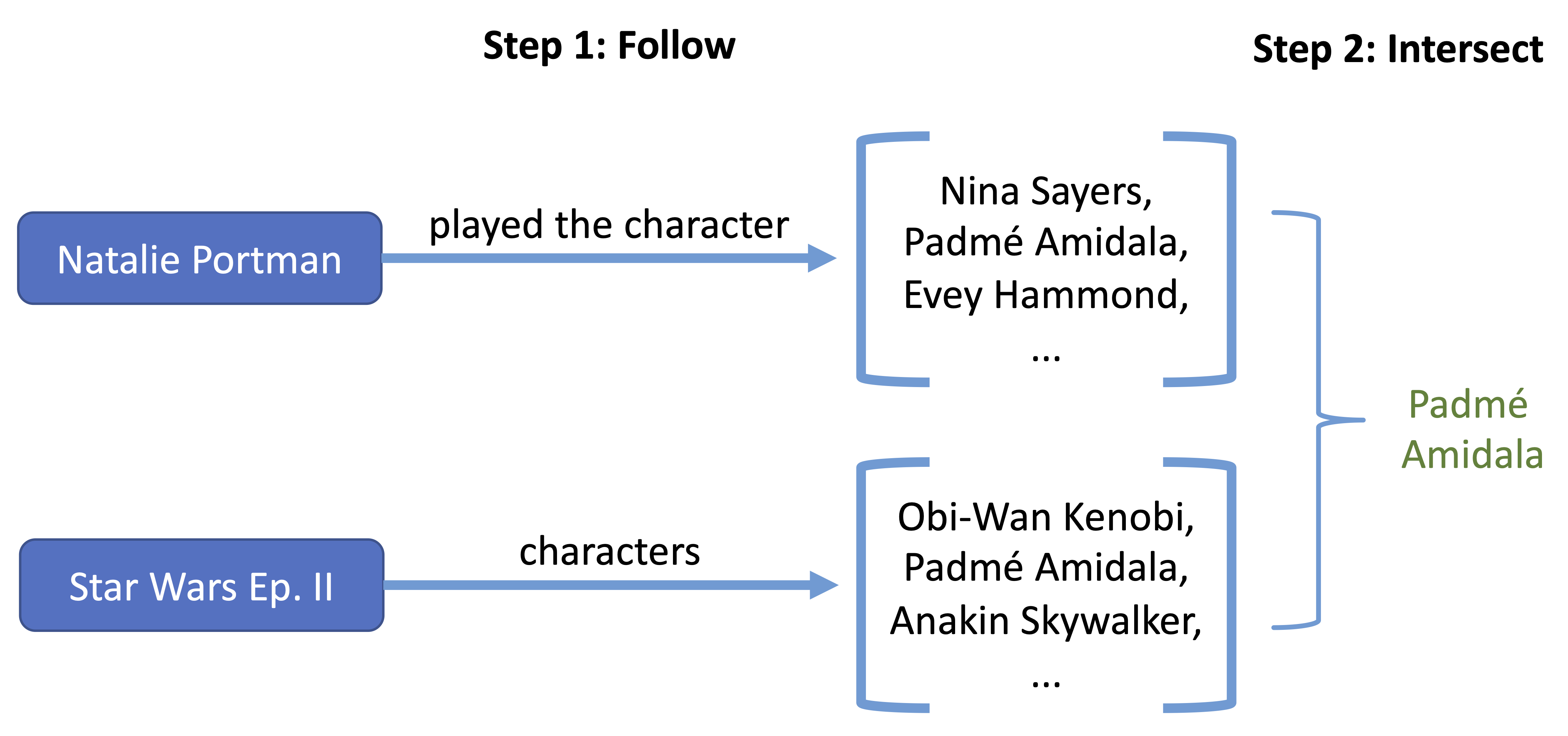}
\caption{To answer “\textit{Who did Natalie Portman play in Star Wars Episode II?}”, we identify all the characters Natalie Portman has played, all the characters in Star Wars Episode II, and intersect the two resulting sets to get to the answer, Padmé Amidala.}
\label{fig:intersection}
\end{figure}

Previous implementations of E2EQA models using differentiable knowledge graphs \cite{cohen2020scalable} have focused on single-entity questions using a relation following operation. For example, to answer “\textit{Where was \underline{Natalie Portman} born?}”, the model could predict a path starting at the Natalie Portman entity and following a \textit{place of birth} relation to the correct answer. 

While this follow operation handles many questions, it often struggles on questions with multiple entities. For example, to answer “\textit{Who did \underline{Natalie Portman} play in \underline{Star Wars Episode II}?}”, it is not enough to identify all the characters Natalie Portman has played, nor all the characters in Star Wars Episode II. Instead, the model needs to find what character Natalie Portman has played that is also a character in Star Wars. This can be solved through \textbf{intersection}. An intersection of two sets \textit{A} and \textit{B} returns all elements in \textit{A} that also appear in \textit{B}. This example is illustrated in Figure \ref{fig:intersection}.

In this paper, we propose to explicitly handle multiple-entity questions in E2EQA by learning intersection in a dynamic multi-hop setting. Our intersection models learn to both follow relations and intersect sets of resulting entities in order to arrive at the correct answer. We find that our models score 73.3\% on WebQuestionsSP and 48.7\% on ComplexWebQuestions, and in particular, improve upon a baseline on questions with multiple entities from 56.3\% to 70.6\% on WebQuestionsSP and 36.8\% to 55.8\% on ComplexWebQuestions.

\section{Related Works}
Traditional approaches to KGQA have used semantic parsing \cite{zelle1996,zettlemoyer2005} to parse natural language into a logical form. Collecting semantic parsing training data can be expensive and is done either manually \cite{dahl1994, finegan-dollak-etal-2018-improving} or using automatic generation \cite{wang-etal-2015-building} which is not always representative of natural questions \cite{herzig-berant-2019-dont}. 

Another line of work in KGQA uses embedding techniques to implicitly infer answers from knowledge graphs. These methods include GRAFT-Net \cite{sun-etal-2018-open}, which uses a graph convolutional network to infer answers from subgraphs, PullNet \cite{sun-etal-2019-pullnet}, which improves GRAFT-Net by learning to retrieve subgraphs, and EmbedKGQA \cite{saxena-etal-2020-improving}, which incorporates knowledge graph embeddings. EmQL \cite{sun2020faithful} is a query embedding method using set operators,  however these operators need to be pretrained for each KB. TransferNet \cite{shi2021transfernet} is a recent model that trains KGQA in a differentiable way, however it stores facts as an \textit{N} x \textit{N} matrix, where \textit{N} is the number of entities, so it runs into scaling issues with larger knowledge graphs.

Our approach to KGQA based on \newcite{cohen2020scalable} has three main advantages:
\begin{itemize}
\item \textbf{Interpretability}: Models based on graph convolutional networks (PullNet, GRAFT-Net) get good performance but have weak interpretability because they do not output intermediate reasoning paths. Our approach outputs intermediate paths as well as probabilities. 
\item \textbf{Scaling}: \newcite{cohen2020scalable} show that differentiable KGs can be distributed across multiple GPUs and scaled horizontally, so that different triple IDs are stored on different GPUs, allowing for scaling to tens of millions of facts. Other methods using embedding techniques (EmbedKGQA, EmQL) or less efficient representations (TransferNet) are more memory intensive and not easily distributed. 
\item \textbf{No retraining for new entities}: Models based on latent representations of entities (EmbedKGQA, EmQL) get state-of-the-art performance, however they need to be retrained whenever a new entity is added to the KG (e.g., a new movie) to learn updated embeddings. Our approach can incorporate new entities easily without affecting trained models.
\end{itemize}

\begin{figure*}[ht]
\includegraphics[width=1\textwidth]{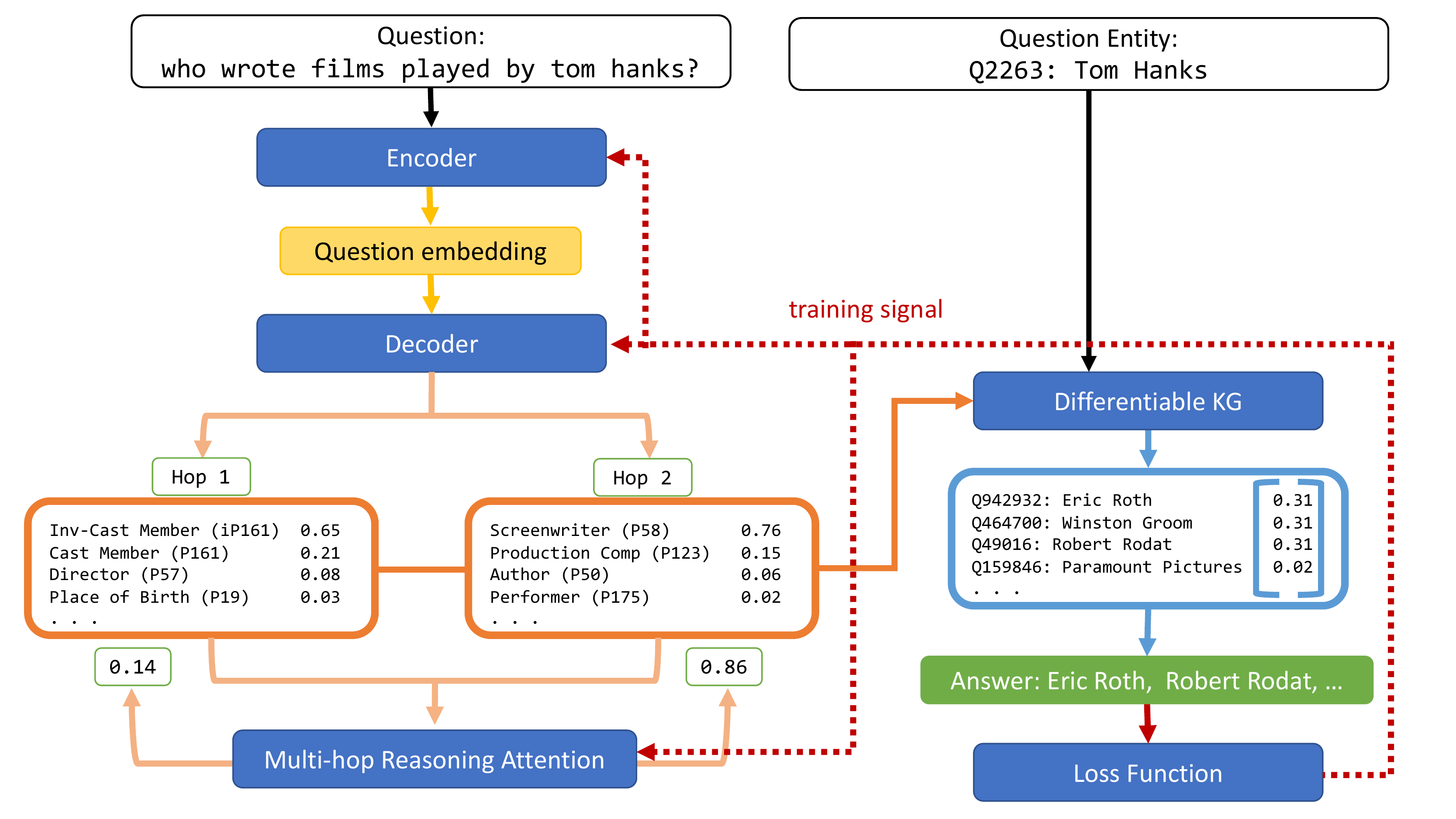}
\caption{A detailed illustration of the Rigel-Baseline model. The encoder encodes the question, the decoder predicts relations, and attention selects the final hop. The entities and relations are followed in the KG to return predicted answers. The loss between the predicted and actual answers is the training signal for the whole model.}
\label{fig:rigel-baseline}
\end{figure*}

\begin{figure*}[ht]
\includegraphics[width=1\textwidth]{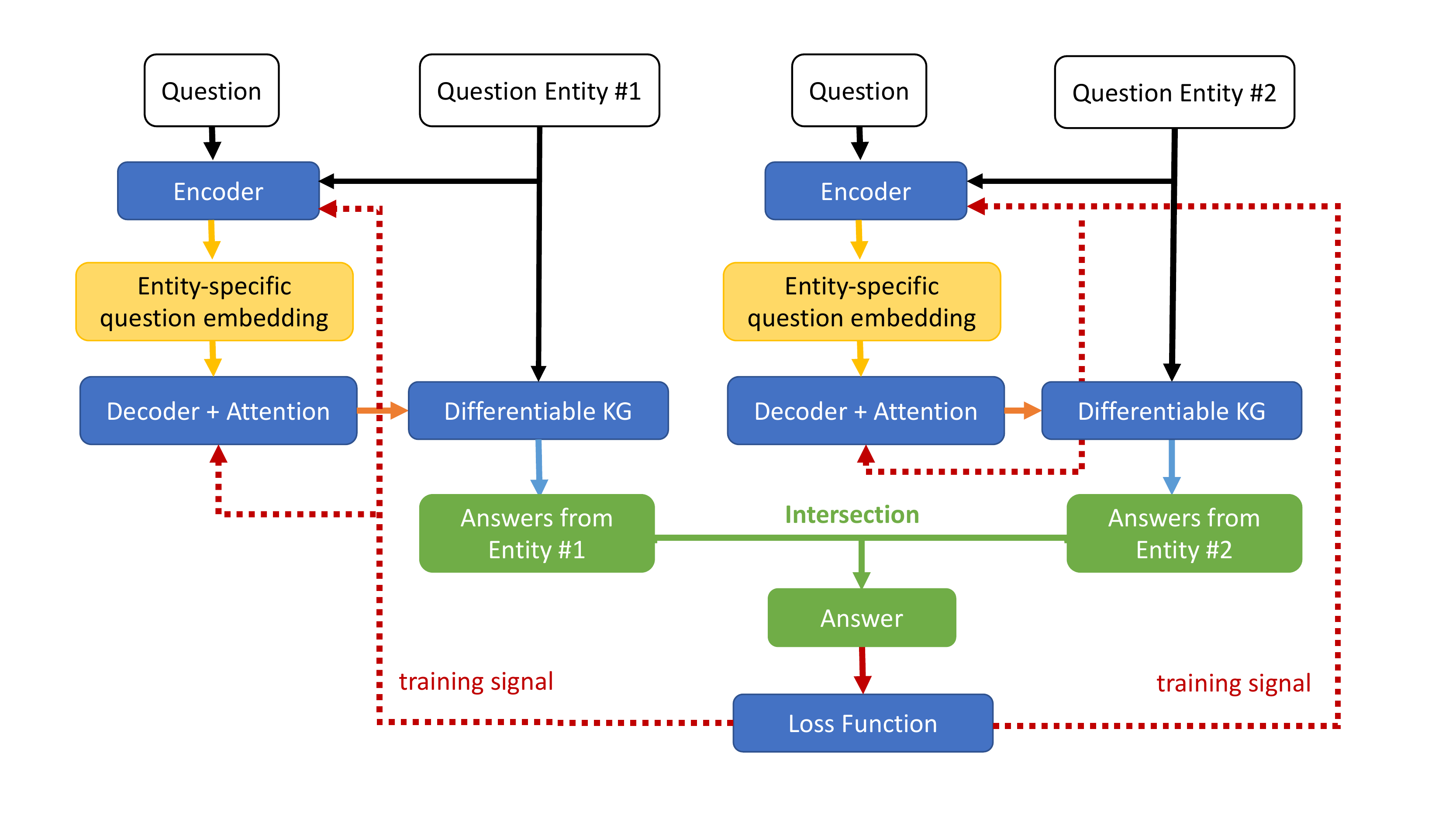}
\caption{An illustration of the Rigel-Intersect model. Given a question with two entities, we run each question entity in parallel to return intermediate answers. These answers are intersected to return the final answer. The loss between the final answer and the actual answer is the training signal for the whole model.}
\label{fig:rigel-intersect}
\end{figure*}

\section{Models}

\subsection{The Baseline Model}

Our baseline model, which we call Rigel-Baseline, is based on differentiable knowledge graphs and the ReifiedKB model \cite{cohen2020scalable}. We provide an overview here but full details can be found in the original paper. 

\subsubsection{Differentiable Knowledge Graphs}
Assume we have a graph:
\begin{equation}\label{eq:graph}
  \mathcal{G} = \bigl\{ (s, p, o) \, | \, s \in E, o \in E, p \in R \bigr\} ,
\end{equation}
where $E$ is the set of entities, $R$ is the set of relations, and $(s, p, o)$ is a triple showing that the relation $p$ holds between a subject entity $s$ and an object entity $o$. To create a differentiable knowledge graph, we represent the set of all triples $T = \{t_i\}_{i=1}^{N_T}, \quad t_i = (s_{s_i}, p_{p_i}, o_{o_i})$ in three matrices: a subject matrix ($M_s$), relation matrix ($M_p$), and object matrix ($M_o$). A triple $(s, p, o)$ is represented across all three matrices at a given index. 
\begin{align*}
  M_s \in \{0, 1\}^{N_T \times N_E}, & \quad M_s(i, j) = \mathbb{I} \bigl( e_j = s_{s_i} \bigr) \\
  M_p \in \{0, 1\}^{N_T \times N_R}, & \quad M_p(i, j) = \mathbb{I} \bigl( p_j = p_{p_i} \bigr) \\
    M_o \in \{0, 1\}^{N_T \times N_E}, & \quad M_o(i, j) = \mathbb{I} \bigl( e_j = o_{o_i} \bigr) 
\end{align*}
Since the knowledge graph is represented as matrices, interacting with the knowledge graph is done with matrix operations. ReifiedKB was implemented with a follow operation: Given an entity vector $\mathbf{x}_{t - 1} \in \mathbb{R}^{N_E}$ at $t - 1$-th time step and a relation vector $\mathbf{r}_{t} \in \mathbb{R}^{N_R}$, the resulting entity vector $\mathbf{x}_t$ is computed by Equation \ref{eq:reified-follow-kgqa} where $\odot$ is element-wise multiplication.
\begin{equation}\label{eq:reified-follow-kgqa}
  \mathbf{x}_t = \text{follow}(\mathbf{x}_{t - 1}, \mathbf{r}_t) = M_o^T ( M_s \mathbf{x}_{t - 1} \odot  M_p \mathbf{r}_t )
\end{equation}

\subsubsection{Model}
The Rigel-Baseline model is composed of an encoder, which encodes the question, and a decoder, which returns a probability distribution over KG relations. The question entities (which, in our experiments, are provided from the datasets) and predicted relations are followed in the differentiable knowledge graph to return predicted answers. Predicted answers are compared to labeled answers, and the loss is used to update the model. Rigel-Baseline is illustrated in Figure \ref{fig:rigel-baseline}.

We make the following key improvements to ReifiedKB. First, we use RoBERTa \cite{liu2019roberta} as our encoder instead of word2vec. Second, ReifiedKB used different methods to determine the correct number of hops in multi-hop questions. We implement an attention mechanism using a hierarchical decoder $W_t^{dec}$, which is learned and a unified approach across datasets. Given a question embedding $\mathbf{h}_q$ and relation vector $\mathbf{r}_t$,  the resulting entity vector $\mathbf{x}_t$ is computed as:
\begin{align}\label{eq:rigel-inference}
  \mathbf{r}_t = & \text{softmax} \Bigl( W_t^{dec} \bigl[ \mathbf{h}_q | \mathbf{r}_{t - 1} | \cdots | \mathbf{r}_{1} \bigr]^T \Bigr) \\
  \mathbf{x}_t = & \text{follow}(\mathbf{x}_{t - 1}, \mathbf{r}_t)
\end{align}
We compute an attention score across all hops with:
\begin{align}\label{eq:rigel-attention}
  c_t = & W_t^{att} \bigl[ \mathbf{h}_q | \mathbf{r}_{t - 1} | \cdots | \mathbf{r}_{1} \bigr]^T \\
  \mathbf{a} = & \text{softmax} ([ c_1, \cdots, c_{T_h} ]) 
\end{align}
and compute the final estimate $\hat{\mathbf{y}}$ as: 
\begin{equation}\label{eq:rigel-answer}
  \hat{\mathbf{y}} = \sum_{t = 1}^{T_h} a_t \mathbf{x}_t .
\end{equation}
Finally, while ReifiedKB used cross-entropy as its loss function, we instead use a multi-label loss function across all entities. This is because the output space in many samples contains multiple entities, so cross-entropy loss is inadequate.
\begin{equation}\label{eq:rigel-loss}
  \mathcal{L}(y, \hat{y}) = \frac{1}{N_E} \sum_{i = 1}^{N_E} y_i \log \hat{y}_i + (1 - y_i) \log (1 - \hat{y}_i)
\end{equation}

\subsection{The Intersection Model}

In order to build a model that can handle multiple entities, we expand Rigel-Baseline with a differentiable intersection operation to create Rigel-Intersect. We define intersection as the element-wise minimum of two vectors. While differentiable intersection has previously been implemented as element-wise multiplication \cite{cohen2019neural}, we prefer to use minimum since it prevents diminishing probabilities. Given two vectors \textit{a} and \textit{b}, the element-wise minimum ($\text{min}_{elem}$) returns a minimum (min) at each index where min($a_n$, $b_n$) will return $a_n$ if $a_n$ $<$ $b_n$, or $b_n$ if $b_n$ $<$ $a_n$. Any element that appears in both vectors returns a non-zero value, and elements that appears in one or neither vector return a 0.
\begin{equation}\label{eq:intersect}
\text{min}_{elem}
\begin{pmatrix}
  \begin{bmatrix}
    a_1 \\
    a_2 \\
    ...\\
    a_n 
  \end{bmatrix}
  \lowcomma
  \begin{bmatrix}
    b_1 \\
    b_2 \\
    ...\\
    b_n 
  \end{bmatrix}
 \end{pmatrix}
  = 
  \begin{bmatrix}
    \text{min}(a_1, b_1) \\
    \text{min}(a_2, b_2) \\
    ...\\
    \text{min}(a_n, b_n) 
  \end{bmatrix}
\end{equation}
Next, we modify the encoder to allow entity-specific question embeddings. The Rigel-Baseline encoder creates one generic question embedding per question, but we may want to follow different relations for each entity. To calculate the question embedding $\mathbf{h_q}$ using an encoder $f_q$, for each question entity, we concatenate the question text $q$ with the entity's mention or canonical name $m$ separated by a separator token [SEP]. We use the embedding at the separator token index ${i_{SEP}}$ as the entity-specific representation of the question. 
\begin{equation}
    \mathbf{h_q} = f_q(q\text{ [SEP] }m)[:, i_{SEP}, :]
\end{equation}

In the decoder, we predict inference chains in parallel for each entity, and follow the entities and relations in the differentiable KG to return intermediate answers. We intersect the two intermediate answers to return the final answer. In multi-hop settings, we weight entities in each vector before intersection based on the attention score. We train using the hyperparameters in Appendix A. Rigel-Intersect is illustrated in Figure \ref{fig:rigel-intersect}.

Our implementation of intersection takes a maximum of two entities per question. We use the first two labeled entities per question and ignore subsequent ones. This works well for our datasets where 94\% of questions contain a maximum of two entities. Given a dataset with more complex questions, we can extend our implementation in the future to an arbitrary number of entities. 

\section{Datasets}

We use two datasets in our experiments:

\textbf{WebQuestionsSP} \cite{yih2016value} is an English question-answering dataset of 4,737 questions (2,792 train, 306 dev, 1,639 test) that are answerable using Freebase \cite{bollacker2008freebase}. During training, we exclude 30 training set questions with no answer, and during evaluation, we exclude 13 test set questions with no answer and count them as failures in our Hits@1 score. All questions are answerable by 1 or 2 hop chains of inference, so we set our models to a maximum of 2 hops. To create a subset of Freebase, we identify all question entities and relations in WebQuestionsSP and build a subgraph containing all facts reachable from 2 hops of the question entities, as done in previous works \cite{cohen2020scalable}. This creates a subgraph of 17.8 million facts, 9.9 million entities, and 670 relations. We create inverse relations for each relation (e.g., \textit{place of birth} returns a person's birthplace; \textit{inv-place of birth} returns people born in that location), for a total of 1,340 relations.

\textbf{ComplexWebQuestions} \cite{talmor2018web} is an extended version of WebQuestionsSP with 34,689 questions (27,649 train, 3,509 dev, 3,531 test) in English requiring complex reasoning. During training, we exclude 163 questions that are missing question or answer entities, and during evaluation, we exclude 21 examples from the test set for the same reasons and count them as failures in the Hits@1 score. We limit our model to 2 hops for training efficiency. To create a subset of Freebase, we identify all question entities and relations in the dataset and build a subgraph containing all facts reachable within 2 hops of the question entities. This results in a subgraph of 43.2 million facts, 17.5 million entities, and 848 relations (1,696 including inverse relations).

\section{Results}

\begin{table}[htb]
\begin{center}
\begin{tabular}{l c c}
\toprule
Model & WQSP & CWQ  \\
\midrule
KVMem \cite{miller-etal-2016-key} & 46.7 & 21.1 \\
GRAFT-Net \cite{sun-etal-2018-open} & 67.8 & 32.8 \\
PullNet \cite{sun-etal-2019-pullnet} & 68.1 & 47.2 \\
ReifiedKB \cite{cohen2020scalable} & 52.7 & -- \\
EmQL \cite{sun2020faithful} & \textbf{75.5} & -- \\
TransferNet \cite{shi2021transfernet} & 71.4 & 48.6 \\
\midrule
Rigel-Baseline (ours) & 69.6 &  39.8 \\
Rigel-Intersect (ours) & 73.3 & \textbf{48.7}  \\
\bottomrule
\end{tabular}
\end{center}
\caption{Comparison of Hits@1 results on WebQuestionsSP (WQSP) and ComplexWebQuestions (CWQ)}
\label{tab:results_compare}
\end{table}

\begin{table}[htb]
\begin{center}
\begin{tabular}{l c c c}
\toprule
Dataset & Baseline & Intersect \\
\midrule
WebQSP & 69.6 &  73.3  \\
WebQSP 1 Entity & 75.8 & 75.3 \\
WebQSP >1 Entity & 56.3 & \textbf{70.6} \\
\midrule
CWQ & 39.8 & 48.7 \\
CWQ 1 Entity & 43.1 &  42.7 \\
CWQ >1 Entity & 36.8 & \textbf{55.8} \\
\bottomrule
\end{tabular}
\end{center}
\caption{Hits@1 breakdown by number of entities for Rigel-Baseline and Rigel-Intersect}
\label{tab:results_breakdown}
\end{table}

Our results are in Tables \ref{tab:results_compare} and \ref{tab:results_breakdown}. Scores are reported as Hits@1, which is the accuracy of the top predicted answer from the model. Table \ref{tab:results_compare} compares our scores to previous models. The aim of our paper is to show an extension of a promising KGQA technique, not to produce state-of-the-art results, but this table shows that Rigel-Intersect is competitive with recent models. Our improved Rigel-Baseline scores higher than ReifiedKB on WebQuestionsSP at 69.6\%, and Rigel-Intersect improves upon that at 73.3\%. On ComplexWebQuestions, Rigel-Baseline scores lower than recent methods at 39.8\%, but Rigel-Intersect gets competitive results with 48.7\%. 

The breakdown of results in Table \ref{tab:results_breakdown} shows that the improved performance of Rigel-Intersect comes from better handling of questions with multiple entities. While Rigel-Baseline and Rigel-Intersect are comparable on questions with one entity, Rigel-Intersect surpasses Rigel-Baseline on questions with more than 1 entity by over 14\% on WebQuestionsSP (56.3\% vs. 70.6\%) and by 19\% on ComplexWebQuestions (36.8\% vs. 55.8\%). Example model outputs are in Appendix C.

Rigel-Baseline is not incapable of handling multiple-entity questions because not all questions require intersection. For example, in "\textit{Who played Jacob Black in Twilight?}", the model can follow \textit{Jacob Black} to the actor, \textit{Taylor Lautner}, without intersecting with \textit{Twilight} because only one actor has played Jacob Black. This is not possible for characters such as James Bond or Batman, who are portrayed by different actors in different movies. Although Rigel-Baseline can spuriously handle multiple-entity questions, Rigel-Intersect uses more accurate inference chains.

\section{Conclusions}

In this paper, we expand an end-to-end question answering model using differentiable knowledge graphs to learn an intersection operation. We show that introducing intersection improves performance on WebQuestionsSP and ComplexWebQuestions. This improvement comes primarily from better handling of questions with multiple entities, which improves by over 14\% on WebQuestionsSP, and by 19\% on ComplexWebQuestions. In future work, we plan to expand our model to more operations, such as union or difference, to continue improving model performance on complex questions.

\bibliography{anthology,custom}
\bibliographystyle{acl_natbib}

\appendix
\section{Model Hyperparameters}

We train Rigel-Baseline and Rigel-Intersect using the hyperparameters shown below. For WebQuestionsSP, we train on a single 16GB GPU. Training completes in approximately 12 hours. ComplexWebQuestions is a larger dataset with a larger knowledge graph, so we train on 4 16GB GPUs, with the knowledge graph distributed across 3 GPUs and the model on the fourth GPU. Training completes in approximately 40 hours.

\begin{table}[hbt!]
    \begin{center}
        \begin{tabular}{l r r}
            \toprule
            Hyperparameter & WQSP & CWQ \\
            \midrule
            Batch Size & 4 & 2\\
            Gradient Accumulation & 32 & 64\\
            Training Steps & 40000 & 80000\\
            Learning Rate & 1e-4 & 1e-4\\
            Max Number of Hops & 2 & 2\\
            \bottomrule
        \end{tabular}
    \end{center}
    \caption{Hyperparameters for training on WebQuestionsSP (WQSP) and ComplexWebQuestions (CWQ)}
    \label{tab:hyperparam}
\end{table}

\section{Validation and Test Performance}
The table below shows the corresponding validation  performances on the dev set for each test result.

\begin{table}[hbt!]
    \begin{center}
        \begin{tabular}{l l r r}
            \toprule
            Dataset & Model & Dev & Test \\
            \midrule
            WQSP & Rigel-Baseline & 72.9 & 69.6\\
            & Rigel-Intersect & 77.1 & 73.3\\
            \midrule
            CWQ & Rigel-Baseline & 40.8 & 39.8\\
            & Rigel-Intersect & 48.1 & 48.7\\
            \bottomrule
        \end{tabular}
    \end{center}
    \caption{Validation and test performance in terms of Hits@1 for WebQuestionsSP (WQSP) and ComplexWebQuestions (CWQ)}
    \label{tab:dev}
\end{table}

\section{Examples}

The table on the following page shows example outputs of Rigel-Baseline and Rigel-Intersect. We only show the top predicted inference chain for each question, but in practice, a probability distribution over all relations is returned. The model also predicts how many hops to take based on an attention score. In our examples, if the model predicts one hop, we show only the top relation from the first hop. If the model predicts two hops, then we show the top relations for both hops.

The first two examples are questions that Rigel-Intersect handles well. In both of these questions, there are multiple probable answers if only one entity is followed (i.e., Russell Wilson attended multiple educational institutions; Michael Keaton has played multiple roles). However only one answer is correct if all entities are considered. The final question is an example where Rigel-Baseline answers correctly even though there are two entities. This is because there is only one winner of the 2014 Eurocup Finals Championship, which Rigel-Baseline can identify without needing to check if the team is from Spain. 

\begin{table*}[b]
\begin{center}
\begin{tabular}{l}
\toprule
\textbf{Q:} What educational institution with The Badger Herald newspaper did Russell Wilson go to? \\
\textbf{A:} University of Wisconsin–Madison\\ \\

\textbf{Rigel-Baseline} \\
Russell Wilson $\rightarrow$ people.person.education $\rightarrow$ education.education.institution \\
$\rightarrow$ Collegiate School \xmark \\ \\

\textbf{Rigel-Intersect} \\
Russell Wilson $\rightarrow$ people.person.education $\rightarrow$ education.education.institution\\
The Badger Herald $\rightarrow$
<inv>-education.educational\_institution.newspaper\\
\textit{Intersection} $\rightarrow$  University of Wisconsin–Madison \cmark \\

\midrule
\textbf{Q:} Who does Michael Keaton play in Cars?\\
\textbf{A:} Chick Hicks\\ \\

\textbf{Rigel-Baseline} \\
Michael Keaton $\rightarrow$ film.actor.film $\rightarrow$ film.performance.character \\
$\rightarrow$ Birdman \xmark \\ \\

\textbf{Rigel-Intersect} \\
Michael Keaton $\rightarrow$ film.actor.film $\rightarrow$ film.performance.character\\
Cars $\rightarrow$ film.film.starring $\rightarrow$
film.performance.character\\
\textit{Intersection} $\rightarrow$ Chick Hicks \cmark \\
\midrule

\textbf{Q:} Which popular sports team in Spain won the 2014 Eurocup Finals Championship?\\
\textbf{A:} Valencia BC \\ \\

\textbf{Rigel-Baseline} \\
2014 Eurocup Finals Championship $\rightarrow$ sports.sports\_championship\_event.champion \\
$\rightarrow$ Valencia BC  \cmark \\ \\

\textbf{Rigel-Intersect} \\
Spain $\rightarrow$ <inv>-sports.sports\_team.location\\
2014 Eurocup Finals Championship $\rightarrow$ sports.sports\_championship\_event.champion\\
\textit{Intersection} $\rightarrow$  Valencia BC  \cmark \\
\bottomrule
\end{tabular}
\end{center}
\caption{Example outputs of Rigel-Baseline and Rigel-Intersect}
\label{tab:examples}
\end{table*}

\end{document}